\documentclass[a4paper,twoside]{article}

\usepackage{epsfig}
\usepackage{subcaption}
\usepackage{calc}
\usepackage{amssymb}
\usepackage{amstext}
\usepackage{amsmath}
\usepackage{amsthm}
\usepackage{multicol}
\usepackage{pslatex}
\usepackage{apalike}
\usepackage{SCITEPRESS}     
\usepackage{booktabs,tabularx}
\usepackage{multirow}
\usepackage{verbatim}
\usepackage{url}

\begin{document}

\title{SparseDet: Towards End-to-End 3D Object Detection}

\author{\authorname{Jianhong Han\sup{1}, Zhaoyi Wan\sup{2}, Zhe Liu\sup{3}, Jie Feng\sup{1} and Bingfeng Zhou\sup{1}}
\affiliation{\sup{1}Wangxuan Institute of Computer Technology, Peking University, Beijing, China}
\affiliation{\sup{2}University of Rochester, Rochester, U.S.A.}
\affiliation{\sup{3}Huazhong University of Science and Technology, Wuhan, China}
\email{\{hanjh, feng\_jie, cczbf\}@pku.edu.cn, i@wanzy.me, zheliu1994@hust.edu.cn}
}

\keywords{3D Deep Learning, Object Detection, Point Clouds, Scene and Object Understanding}

\abstract{In this paper, we propose \textit{SparseDet} for end-to-end 3D object detection from point cloud.
  Existing works on 3D object detection rely on dense object candidates over all locations in a 3D or 2D grid following the mainstream methods for object detection in 2D images.
  However, this dense paradigm requires expertise in data to fulfill the gap between label and detection.
  As a new detection paradigm, SparseDet maintains a fixed set of learnable proposals to represent latent candidates and directly perform classification and localization for 3D objects through stacked transformers.
  It demonstrates that effective 3D object detection can be achieved \textbf{with none of} post-processing such as redundant removal and non-maximum suppression.
  With a properly designed network, SparseDet achieves highly competitive detection accuracy while running with a more efficient speed of 34.5 FPS.
  We believe this end-to-end paradigm of SparseDet will inspire new thinking on the sparsity of 3D object detection.}

\onecolumn \maketitle \normalsize \setcounter{footnote}{0} \vfill

\section{\uppercase{Introduction}}
\label{sec:introduction}

3D Object detection is a technology used to identify the category and location of an object of interest in a scene.
As one of the fundamental topics in the field of computer vision, it plays a pivotal role in applications for a wide range of scenarios, e.g., autonomous driving and augmented reality.
With the advancement of 3D sensors, considerable research attention has been paid to 3D object detection from point cloud in recent years.
Due to its representation power from a low-resolution resampling of the real 3D world geometry, point cloud scanned from LiDAR depth sensors achieves remarkable success in this field.
Meanwhile, the sparsity and irregularity of point cloud data pose great challenges to accurate and robust 3D object detection.

For the task of 3D object detection from point cloud, we aim at representing an object with an oriented 3D bounding box around it given the point cloud of a scene as input.
To this end, researchers in 3D object detection have proposed various effective methods.
Mainstream 3D object detectors perform dense regression from a large set of coarse proposals or anchors in a rigid spatial grid.
Despite the success that this paradigm has achieved, it is limited in several aspects. (1) The dense prediction requires and is sensitive to \textbf{post-processing procedures}, e.g., non-maximum suppression (NMS), to form the final detection. (2) \textbf{Hyper-parameters} such as sizes and aspect ratios of anchors must be carefully tuned for effective detection. (3) The gap between intermediate representation and final detection makes empirical \textbf{label assignment} strategies critical to detectors.

\begin{table}[tp]
\centering
\caption{Directly applying sparse prediction in a 3D object detection algorithm VoxelNet~\cite{zhou2018voxelnet} suffers from significant degradation of performance. SparseDet makes the first sparse 3D object detector and demonstrates effectiveness and efficiency.}
\begin{tabularx}{1.0\linewidth}{@{}l*{6}X@{}}
\toprule
Method        & Easy   &  Mod.  & Hard &  mAP  \\ \midrule
VoxelNet     &  81.98  &    65.46    & 62.85 &   70.10  \\
sparse VoxelNet & 61.17 &	56.73 &	56.61  & 58.17  \\
SparseDet     &   88.36	&   78.45   & 77.03 &	81.28      \\
   \bottomrule
\end{tabularx}
\label{tab:vanilla}
\end{table}

\begin{figure*}[!tp]
    \centering
    \includegraphics[width=0.9\textwidth]{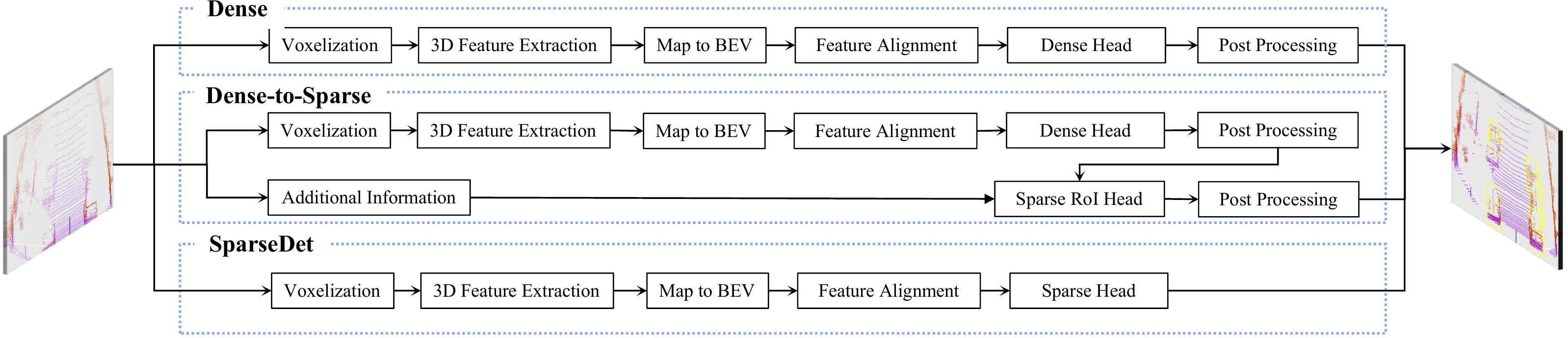}
    \caption{Comparison with mainstream 3D object detection paradigms.
    Distinguishable from conventional dense and dense-to-sparse approaches, SparseDet performs end-to-end detection without post-processing.
    }
    \label{fig:arch}
\end{figure*}

Drawing inspiration from object detection in 2D images, we explore end-to-end 3D object detection in a sparse manner in this paper.
To fit into the nature of sparsity in point cloud, we propose to apply sparse prediction for detection, thus making an end-to-end paradigm for 3D object detection.
Although end-to-end detection is proven feasible in 2D images~\cite{carion2020end}, an effective design of sparse detector for 3D object in point cloud is non-trivial due to the challenges and differences in tasks.
As shown in Tab.~\ref{tab:vanilla}, intuitively applying sparse prediction in 3D object detection thus suffers from significant degradation of detection accuracy.
For the purpose of an effective end-to-end 3D object detector, we closely investigate the characteristics of detection in point cloud and accordingly design an algorithm that detects 3D objects through sparse prediction, namely \textit{SparseDet}.

SparseDet bears several major differences from previous practices in 3D object detection. Firstly, it discards the enumeration through the voxel grid to generate dense proposals. Instead, it keeps the sparsity by maintaining a set of learnable proposals that are concentrated on the most possible regions in the point cloud. Secondly, it produces object detection without intermediate representation based on label assignment. Thus, requirements of any post-processing such as NMS are eliminated. Thirdly, it relies on none of pre-defined anchors but adaptively learns proposal initialization from a statistics perspective.
We conduct extensive experiments to demonstrate the effectiveness of the proposed SparseDet. Specifically, state-of-the-art detection accuracy performance of 81.3\% 3D mAP and efficiency performance of 34.5 FPS is achieved in KITTI.
Moreover, exhaustive ablation studies are provided to reveal the insights of our model design. 
What we prefer more about the proposed sparse paradigm is it demonstrates promising robustness, which is crucial to the applicable value for a 3D object detector (see Sec.~\ref{sec:roubstness} for details).

Our contribution in this paper can be summarized as follows: 
\begin{itemize}
    \item We are the first to conduct an end-to-end 3D object detector in point cloud by applying a sparse prediction paradigm.
    \item In the context of sparse prediction, we closely investigate the concepts for model design, including a proper feature extractor and targeted loss functions, which provide guidance for effective practice of end-to-end 3D object detection.
    \item Extensive experiments validate the effectiveness of the proposed method: it is a trade-off between accuracy and efficiency, and demonstrates superiority in robustness.
\end{itemize}

\section{\uppercase{Related Work}}

According to the manner of generating boxes, the existing methods for 3D object detection are mainly divided into two categories, including the dense and dense-to-sparse methods. As shown in Fig.~\ref{fig:arch}.

\vspace{1ex}\noindent\textbf{Dense 3D Object Detection.} 
The early dense 3D detectors mainly adopted the anchor-based design, which requires manually setting anchors for each category object. A typical representative method is VoxelNet~\cite{zhou2018voxelnet}, which first extracts the voxel features by a Voxel Feature Encoding layer. Then, a region proposal network~(RPN) takes the voxel feature as input and generates the dense 3D boxes on the detection head. SECOND~\cite{yan2018second} utilizes a more efficient sparse convolution operation to learn voxel feature representation for improving VoxelNet. Based on SECOND, Pointpillars~\cite{lang2019pointpillars} divides the point cloud as a special voxel called pillars to further improve the running speed. However, the pillar-based manner inevitably loses some important context and spatial information. Towards this goal, TANet~\cite{liu2020tanet} proposes a TA module to obtain more robust and distinguishable features.
Recently, anchor-free 3D detector CenterPoint~\cite{yin2020center} regards objects as points and discards the hand-crafted anchor boxes.
Moreover, all these methods need NMS post-processing operation through setting a suitable score threshold to filter out the redundant boxes in inference. 
Which is a common disadvantage of the dense 3D detection paradigm.

\vspace{1ex}\noindent\textbf{Dense-to-Sparse 3D Object Detection.} 
F-PointNet~\cite{qi2018frustum}, Frustum-ConvNet~\cite{wang2019frustum} and SIFRNet~\cite{zhao20193d} first generate a set of 2D region proposals on a front view image with the help of offline strong 2D detectors~\cite{liu2016ssd,redmon2018yolov3}. Then, each 2D proposal is converted into a 3D viewing frustum in point cloud. Finally, PointNets~\cite{qi2017pointnet,qi2017pointnet++} is applied to estimate the final 3D boxes for each 3D viewing frustum. However, the manner is limited by the performance of 2D detectors.
Instead of F-PointNet~\cite{qi2018frustum},
~\cite{shi2019pointrcnn,yang2019std,shi2020points,chen2019fast,huang2020epnet,chen2017multi,ku2018joint,deng2020voxel} 
directly achieve the online dense-to-sparse paradigm. Specifically, these methods first produce a large number of boxes via a RPN. Then, to provide the high-quality boxes for the RCNN stage, the NMS is also a necessary operation during the process from the first stage to the second stage. Finally, the RCNN further estimates the final 3D boxes via refining the sparse boxes from the RPN stage. These approaches can usually obtain high performance but a slower running speed compared with the dense 3D object detectors. In contrast, our SparseDet not only abandons some tedious manual operation~(e.g. \emph{anchors and NMS} ), but also can achieve comparable results with a high running speed.

\vspace{1ex}\noindent\textbf{Sparse 2D Object Detection.} 
Recently, with the widespread application of transformers in computer vision, sparse object detection methods have attracted more and more attention. DETR~\cite{carion2020end} builds the first fully end-to-end 2D object detector, which not only eliminates the anchor design and NMS operation but also achieves comparable results with the existing detectors. Based on DETR, D-DETR~\cite{zhu2020deformable} mitigates the slow convergence and high complexity issues of DETR through the proposed deformable attention module. Sparse R-CNN~\cite{peize2020sparse} completely remove to object candidates design and achieve better performance and faster convergence through the learnable proposal boxes and their corresponding features than DETR and D-DETR. Motivated by these sparse methods, we propose SparseDet in this paper, which aims to provide a strong and simple baseline serving for the 3D object detection community.

\section{\uppercase{SparseDet for 3D Object Detection}}\label{sec:method}

In this section, we describe the novel SparseDet to explore the feasibility of a sparse detector from 3D point cloud. The overall architecture is illustrated in Fig.~\ref{fig:arch}.
The paradigm of SparseDet can be roughly divided into three main components: a voxel-based feature extractor, a feature alignment module that refines the extracted features, and a detection head performing detection in a sparse manner. In the following, we introduce each of the components of SparseDet and the loss functions for the objective of training in detail.

\subsection{Voxel Feature Extraction}

Compared with two-stage point-based methods~\cite{shi2019pointrcnn,huang2020epnet}, voxel-based approaches~\cite{lang2019pointpillars,yan2018second} occupy a higher efficiency. Therefore, our method focus on voxel-based methods. Specifically,
We first encode the point cloud into a regular volume grid, namely voxels, thus effectively compressing the representation with aggregated features.
However, point cloud is of uneven distribution since it is collected by LiDAR sensors distant from targets. 
Therefore, the number of points to be represented in each voxel varies dramatically.
To avoid being overwhelmed by voxels with overly dense points, we limit the maximum number of points as $T$ in each voxel.
The neutralization is implemented by random sampling inside these voxels.

More formally, given a point cloud in the range of $L \times W \times H$ meters and voxel size of $D_x$, $D_y$, $D_z$, the discrete voxel feature grid with the shape of $S_x \times S_y \times S_z$ is obtained, where
\begin{equation}
    \begin{split}
        S_x = \lceil\frac{L}{D_x}\rceil,
        S_y = \lceil\frac{W}{D_y}\rceil,
        S_z = \lceil\frac{H}{D_z}\rceil.
    \end{split}
\end{equation}

After voxelization, non-empty voxels are maintained and encoded as the mean of point-wise features of all inside points.Then the voxel features are fed into a stack of 3D sparse convolution layers~\cite{graham2014spatially,graham2015sparse,graham2017submanifold,graham20183d,yan2018second} to extract rich features from the voxel representation.

\subsection{Feature Alignment}

We avoid directly exploiting the 3D voxel features for prediction, which bring high computational cost.
Instead, the features in 3D are collapsed into a fixed view, e.g., Birds' Eye View (BEV), to align features along the $Z$-axis.
The conventional BEV feature alignment involves a downsample-and-upsample schema: a sub-network that reduces the resolution with stacked 2D convolutions and a multi-scale feature fusion sub-network.

Therefore, we introduce the Feature Pyramid Network (FPN)~\cite{lin2017feature} and Pyramid Sampling Aggregation (PSA)~\cite{liu2020tanet} modules as the basis for our feature alignment. FPN is designed to solve the problem of scale variance of which objects emerge in images. Although the scale of akin objects is consistent in point cloud due to the differences in sensors, point sparsity and reflection strength decrease for objects far from the sensor.
Therefore, sparse 3D object detection benefits from a pyramid model structure to capture information from different scales.
Given the BEV features where the view is fixed, we obtain features with different solutions. Specifically, $2x$, $4x$, and $8x$ downsampled features in the top-down order are produced by three 4-layers convolution modules, whose number of filters is 64, 128, and 256, respectively. 
The outputs of these blocks are denoted as $B_1$, $B_2$ and $B_3$.

Then, three feature pyramids of each $B-$feature are achieved through down-sampling and up-sampling.
Specifically, $B_1$ obtains its feature pyramids with itself and $2x$ and $4x$ down-sampled features.
$B_2$ obtains its feature pyramids with its $2x$ up-sampled feature, itself, and its $2x$ down-sampled feature.
Similarly, $B_3$ obtains its feature pyramids with two up-sampling operations based on itself. Now, each pyramid will have the same shape features. Laterally concatenate the same size features of each pyramid to get cross-layer features. Followed by a fusion block convolution layer, the new feature pyramid with multi-level information is fed into a FPN-like structure inversely. Namely, the down-top interpolated feature is added to the last hierarchy lateral feature. Finally, we obtain an informative feature list, which will be delivered to the sparse detection head.

\subsection{Sparse 3D Detection Head }

Recently, the attention mechanism has achieved good results in object detection in 2D images. DETR~\cite{carion2020end} uses a sparse set of object queries, to interact with the global image feature. Features of each position in the image can obtain information of other positions, predictions without any hand-crafted components are realized. Subsequently, DETR~\cite{carion2020end} is improved by Deformable-DETR~\cite{zhu2020deformable} through reducing the range of feature search to facilitate the convergence process. Sparse R-CNN~\cite{peize2020sparse} uses a small number of boxes with features as the learning region proposal. Via the self-attention of these proposal features and their interaction with RoI features, it achieves pleasing object feature learning.

\begin{figure}[h]
    \centering
    \includegraphics[width=0.9\columnwidth]{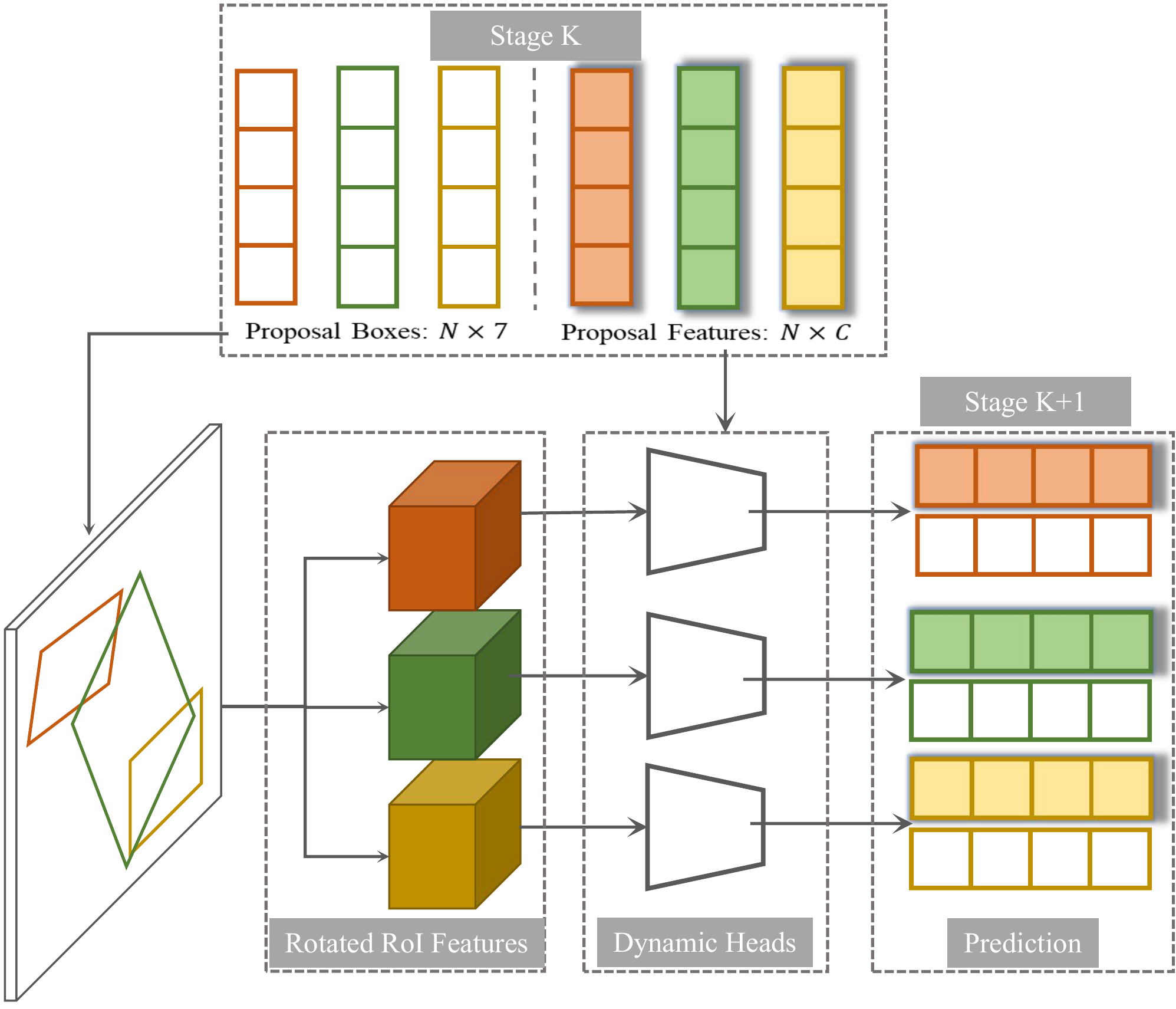}
    \caption{Model design of the sparse detection head, where $K$ represents the index number of stages. The learnable proposals, identified by 3D boxes and feature vectors, interact with the stacked dynamic heads to be refined. Note the proposals produced by the last stage of detection head directly perform as the final detection. Dynamic heads and prediction layers are \textbf{not} shared by proposals.}
    \label{fig:detection-head}
\end{figure}

The model design of our sparse detection head is illustrated in Fig.~\ref{fig:detection-head}.
Inspired by Sparse R-CNN~\cite{peize2020sparse}, we use a small set of learnable bounding boxes with learnable features as object candidates. These sparse candidates are used to extract the feature of Region of Interest (RoI), which avoids hundreds of thousands of candidates from prior anchors or RPN.
Specifically, we use $N$ (e.g., 100) learnable bounding boxes. Each box is represented as the 7-D parameters $(c_x,c_y,c_z,h,w,l,\theta)$, including its center $(c_x,c_y,c_z)$, size $(w, l, h)$, and orientation $\theta$ that indicates the heading angle around the up-axis. However, the 3D bounding box is merely a rough representation of the object, lacking abundant semantic and geometric information. Thus the corresponding learnable proposal features of each box are proposed. Each such feature is represented as a slightly higher dimension latent vector to capture more characteristics of the object. 
The proposal boxes are initialized as the whole point cloud size with no rotation for a larger search space. The proposal features are randomly initialized.

The subsequent learning process will gradually narrow the scope of these boxes until they match the corresponding target. This initialization is effective and reasonable. The parameters of proposal boxes and proposal features will be optimized together with other parameters in the whole network. In fact, what these features eventually learn is the statistical characteristics of the objects that may appear throughout the data.

Then, given the proposal boxes, we can obtain the orientated BEV 2D bounding boxes by discarding the location and scale in the vertical direction. Each BEV box can extract its corresponding feature from the feature map by utilizing the Rotated RoIAlign~\cite{he2017mask} operation. So far, we have two types of features: the proposal features and the RoI features. For $N$ proposal boxes, there are $N$ proposal features, and $N \times S \times S$ RoI features, in which $S$ is the pooling resolution. First, a multi-head attention module followed by a LayerNorm layer is applied to the proposal features to reason about the relations between objects. Then, following Sparse R-CNN, each RoI feature will interact with the corresponding proposal feature to filter out ineffective bins through an exclusive Dynamic Instance Interactive Module. 
The Dynamic Instance Interactive Module is mainly used for making the RoI features and the proposal features communicate with each other to get more discriminate features. Specifically, it is achieved by consecutive $1 \times 1$ convolution with LayerNorm and ReLU activation function.

The output of the Dynamic Instance Interactive Module will serve as the final object features, which are used to compute the classification predictions and 7-D 3D bounding box regression predictions through two Multi Layer Perception (MLP)  with ReLU activation branches.
The regression branch outputs a vector $\boldsymbol{\Delta}=\left(\Delta_{x}, \Delta_{y}, \Delta_{z}, \Delta_{w}, \Delta_{l}, \Delta_{h}, \Delta_{\theta}\right) \in \mathbb{R}^{7}$ that represents the residue from 3D proposal boxes to the ground truth boxes
following predecessors ~\cite{zhou2018voxelnet,yan2018second,lang2018pointpillars,shi2019pointrcnn,liu2020tanet}
:
\begin{equation}
\begin{aligned}
\Delta_{x} &=\frac{x_{g}-x_{p}}{d_{p}}, \Delta_{y}=\frac{y_{g}-y_{p}}{d_{p}}, \Delta_{z}=\frac{z_{g}-z_{p}}{h_{p}} \\
\Delta_{w} &=\log \left(\frac{w_{g}}{w_{p}}\right), \Delta_{l}=\log \left(\frac{l_{g}}{l_{p}}\right), \Delta_{h}=\log \left(\frac{h_{g}}{h_{p}}\right) \\
\Delta_{\theta} &=\theta_{g}-\theta_{p}
\end{aligned}
\end{equation},
where $d_{p}=\sqrt{\left(w_{p}\right)^{2}+\left(l_{p}\right)^{2}}$. Then the residue vector will be decoded to the last prediction box to compose the new prediction.

The proposed Sparse head is stacked several times to perform an iterative structure. The object boxes prediction and object features of the previous stage are fed into the next stage to serve as the proposal boxes and proposal features. Each stacked head is supervised to optimize the proposal boxes and features at different stages.

\subsection{Loss Function}\label{sec:loss}

Current 3D object detectors usually produce thousands of candidates, which can lead to a large number of near duplications. So the NMS post-processing is applied to screen out the fairly good results which exceed a certain score threshold or IoU threshold. There is no method to predict the ultimate objects directly in 3D object detection. While SparseDet is post-processing-free for it aims to predict the final result from the beginning.

When calculating loss, there are two stages. One is the matching stage between the fixed-size set of $N$ predictions and $M$ ground truth objects, where $N$ is set to be significantly larger than the typical number of objects in a point cloud. And the second stage is the optimization of the matched $M$ predictions and ground truth pairs. Unlike the predecessors, our sparse method does not have the concept of anchor, so all losses are directly compared in the prediction results and ground truth.

The $N-M$ matching aims to filter out $M$ competitive candidates from all the $N$ predictions, we follow the bipartite matching loss approach~\cite{carion2020end,zhu2020deformable,peize2020sparse,yang2019learning,stewart2016end} based on the Hungarian algorithm~\cite{kuhn1955hungarian} to compute the pair-wise matching cost. 
Let us denote by 
$y=\left\{y_{i}\right\}_{i=1}^{M}$
the ground truth set of objects, and 
$\hat{y}=\left\{\hat{y}_{i}\right\}_{i=1}^{N}$
the set of N predictions. Where $ N \textgreater M$.
The matching cost is defined as follows:
\begin{equation}
\mathcal{C}=\underset{i \in M, j \in N}{\arg \min }\mathcal{L}_{\operatorname{match}}\left(\hat{y}_{i}, y_{j}\right)
\end{equation},
in which the pair-wise loss is computed as:
\begin{equation}
\mathcal{L}_{\operatorname{match}}=\lambda_{c l s} \cdot \mathcal{L}_{c l s}+\lambda_{L 1} \cdot \mathcal{L}_{L 1}+\lambda_{IoU} \cdot \mathcal{L}_{AA\_BEV\_IoU}
\end{equation}
$\lambda_{c l s}$, $\lambda_{L 1}$ and $\lambda_{IoU}$ are coefficients of each component. $\mathcal{L}_{c l s}$ is focal loss\cite{lin2017focal} of predicted classifications and ground truth category labels.

As for regression loss, we find it is crucial in 3D sparse detection. It almost determines the performance of the detector. For details, please refer to our ablation in Sec.\ref{sec:ablation}. While general 3D detectors only use $L1$ loss to constrain the regression, we find it does not work in 3D sparse detection. The reason may be that sparse proposal is difficult to find the corresponding target merely relying on $L1$ loss. Thus we adopt the union of $L1$ loss and $IoU$ loss following\cite{carion2020end,zhu2020deformable,peize2020sparse}.
$\mathcal{L}_{L 1}$ is $L1$ loss between the normalized predicted box and ground truth box, which has two parts as following:
\begin{equation}
\label{func:l1}
\mathcal{L}_{L 1}=\mathcal{L}_{r e g\_\theta}+\mathcal{L}_{r e g\_{o t h e r}}
\end{equation},
$\mathcal{L}_{r e g\_{o t h e r}}$is the regression loss for 3D center location and box size. $\mathcal{L}_{r e g\_\theta}$ is the regression loss of $\theta$. We find that the angle has a great influence on matching during the research. A proper angle matching method plays a big role. Here Sine-Error loss is adopted following ~\cite{yan2018second} to solve the adversarial example problem between orientations of 0 and $\pi$.

\begin{figure*}[ht]
\begin{center}
  \includegraphics[width=0.96\linewidth]{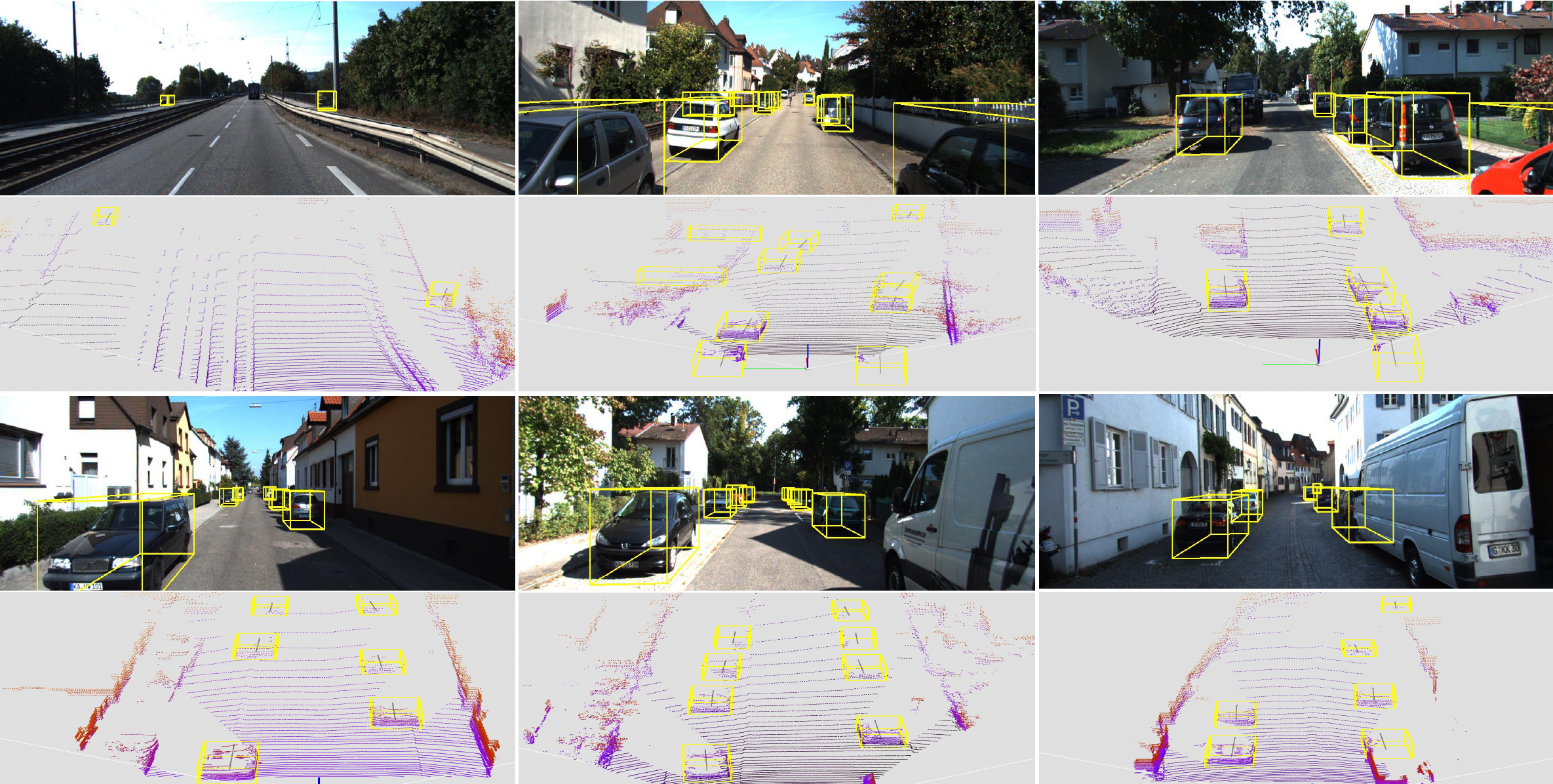}
\end{center}
\caption{Qualitative results on the KITTI dataset. 
For each sample, the upper part is the image for just visualization and the lower part is the point cloud corresponding to the image perspective. The detected objects are shown with yellow 3D bounding boxes.}
\label{visualization}
\end{figure*}

\begin{table*} 
\centering
\caption{Evaluation comparison with state-of-the-art on KITTI validation set. As a detector in the family of voxel-based modality, SparseDet achieves clear and consistent improvements over akin methods.
Theoretically, point-based methods can be augmented with the sparse detection paradigm of SparseDet, although we choose voxel-based methods as our baseline in consideration of inference efficiency. Data of other methods comes from their corresponding paper.}
\begin{tabularx}{\linewidth}{l|c|XXX|XXX|c|c|c} \toprule
\multirow{2}{*}{ Method } & \multirow{2}{*}{ Modality } & \multicolumn{3}{c}{ $AP_{3D}$ $(IoU=0.7)$ } & \multicolumn{3}{|c|} { $AP_{BEV}$ $(IoU=0.7)$ }& \multirow{2}{*}{ FPS }& \multirow{2}{*}{ Param. }& \multirow{2}{*}{ FLOPs } \\
                                &                            &  Mod. &Hard & mAP               &    Mod. & Hard & mAP                &   &                   & \\ \midrule
MV3D   & RGB \& LiDAR  &62.68  & 56.56 & 63.51  &$78.10$ & $76.67$ & 80.44 & 2.78 &-&-\\
ContFuse   & RGB \& LiDAR  & $73.25$& $67.81$ &75.79  &$87.34$& $82.43$ & 88.40 & 16.67 & -&- \\
AVOD-FPN  & RGB \& LiDAR  & $74.44$ & $68.65$ & 75.83&- & -& -& 10.00 &- &- \\
F-PointNet  & RGB \& LiDAR  & $70.92$ & $63.65$ & 72.78& $84.02$  & $76.44$ & 82.87 & 5.88 &-&-\\ \midrule
VoxelNet & Voxel-based  & $65.46$ & $62.85$ & 70.07  & $84.81$ & $78.57$ & 84.33 & 4.35 & -&- \\
SECOND & Voxel-based  & $76.48$ & $69.10$ & 77.67 & $87.07$  & $79.66$ & 85.56 & 20.00 & 5.33M & 76.70G \\
TANet & Voxel-based & 77.85  & 75.62& 80.56 &- &- &- &28.78 &- &-\\ \midrule
PointRCNN  & Point-based & $78.63$ & $77.38$ & 81.63 & $87.89$  & $85.51$ & 87.87 & 8.33 & 4.04M & 27.38G\\
Part-$A^{2}$-f  & Point-based   & $78.96$& $78.36$ & 81.93&  $88.05$  & $85.85$ & 88.04 & 12.50 & 59.23M&-\\
Part-$A^{2}$-a   & Point-based & $79.47$ & $78.54$ & 82.49  & 88.61 & 87.31 & 88.78 & 12.50 & 63.81M &-\\ \midrule
SparseDet  & Voxel-based 	 & 78.45 & 77.03 & 81.28  & 88.02 & 86.49 & 88.11 & 34.48 & 25.10M	& 37.98G\\ \bottomrule
\end{tabularx}
\label{tab:sota}
\end{table*}

In terms of $IoU$ loss, it is used to constrain the overlapping degree between boxes. For 3D oriented box matching, it is natural to think of 3D rotated IoU between boxes. But it is difficult to optimize all dimensions together, which makes the detector confused. So we design a two-stage IoU matching style to make it easier to learn the relationship between boxes.

At the matching stage, Axis-Aligned (AA) BEV 2D bounding box is used to compute the $IoU$ loss of prediction and ground truth box, denoted as $\mathcal{L}_{AA\_BEV\_IoU}$. Axis-Aligned means turning the box to X-axis or Y-axis according to the origin orientation angle $\theta$. BEV means the position and scale of the bounding box on the Z-axis (height axis) are not considered.
The purpose of these operations is to reduce the difficulty of matching orientation and height with other parameters at the same time, making it easier to catch the rough shape. More accurate box matching will be executed at the optimization stage of matched pairs.

The training loss is only performed on matched pairs. Which is almost the same as matching cost, except the $IoU$ loss adopts Rotated 3D $DIoU$ loss denoted as $\mathcal{L}_{R\_3D\_DIoU}$. The Rotated 3D $DIoU$ loss~\cite{zheng2020distance} encode the orientation and height information to make the results more precise:
\begin{equation}
\mathcal{L}=\lambda_{c l s} \cdot \mathcal{L}_{c l s}+\lambda_{L 1} \cdot \mathcal{L}_{L 1}+\lambda_{IoU} \cdot \mathcal{L}_{R\_3D\_DIoU}
\end{equation}

The final loss is the sum of all matched pairs normalized by the number of ground truth objects inside the batch. 

\section{\uppercase{Experiments}}

\begin{table*}[t]
\centering
\caption{Evaluation results on KITTI Car detection under different IoU thresholds. In combination with sparse prediction, SparseDet demonstrates significant advantages in accurate localization. Data of other methods come from \cite{zhou2019IoU}.}
\begin{tabularx}{\textwidth}{l|l|XXX|XXX|XXX}  \toprule
\multirow{2}{*}{ Metric } & \multirow{2}{*}{ Method } & \multicolumn{3}{c|}{ $IoU=0.7$ } & \multicolumn{3}{c|}{ $IoU=0.75$ } & \multicolumn{3}{c} { $IoU=0.8$ } \\ \cline{3-11}
                          &                           & Easy & Mod. & Hard & Easy & Mod. & Hard & Easy & Mod. & Hard \\ \midrule

\multirow{4}{*}{$AP_{3D}$}      & PointPillars\cite{lang2019pointpillars}  & $87.29$ & $76.99$ & $70.84$ & $72.39$ & $62.73$ & $56.40$ & $47.23$ & $40.89$ & $36.31$ \\
                         & \textit{+3D IoU Loss}\cite{zhou2019IoU} & $87.88$ & $77.92$ & $75.70$ & $76.18$ & $65.83$ & $62.12$ & $57.82$ & $45.03$ & $42.95$ \\
                         & \textit{+3D GIoU Loss}\cite{zhou2019IoU} & $\mathbf{8 8 . 4 3}$ & $7 8 . 1 5$ & $7 6 . 3 4$ &76.93&66.36 &63.68 & $56.36$ & $44.43$ & $42.72$ \\
                         & Ours & 88.36&	$\mathbf{78.45}$&	$\mathbf{77.03}$&	$\mathbf{77.58}$&	$\mathbf{68.02}$&$\mathbf{65.61}$&		$\mathbf{63.38}$&	$\mathbf{52.56}$&	$\mathbf{47.22}$\\ \midrule

\multirow{4}{*}{$AP_{BEV}$}     & PointRCNN\cite{shi2019pointrcnn} & $88.14$ & $77.58$ & $75.36$ & $73.27$ & $63.54$ & $61.08$ & $44.21$ & $38.88$ & $34.62$ \\
                         & \textit{+3D IoU Loss}\cite{zhou2019IoU} & $88.83$ & $78.80$ & 78.18 & $77.42$ & $67.83$ & 66.85 & $58.22$ & $49.09$ & $45.38$ \\
                         & \textit{+3D GIoU Loss}\cite{zhou2019IoU} & 88.84 & 78.85 & $78.15$ & 77.47 & 67.98 & 67.18& 59.80 & 51.25 & 46.50 \\
                         & Ours & $\mathbf{89.62}$ &	$\mathbf{88.02}$ &	$\mathbf{86.49}$	&	$\mathbf{88.62}$ &	$\mathbf{84.09}$&	$\mathbf{78.83}$&		$\mathbf{78.41}$&	$\mathbf{75.03}$&	$\mathbf{70.83}$ \\ \bottomrule
\end{tabularx}
\label{tab:ap}
\end{table*}

\begin{table}[t]
\centering
\caption{Robustness evaluation using metrics from ~\cite{liu2020tanet}. SparseDet degrades slower with the number of noise points growing. }
\begin{tabularx}{\columnwidth}{l|c|ccc}
\hline \multirow{2}{*}{ Method } & \multirow{2}{*}{+noise} & \multicolumn{3}{c} { $AP_{3D}$ $(IoU=0.7)$ } \\ \cline { 3 - 5 } 
& & Mod. & Hard & mAP \\ \midrule
PointRCNN     & 0   & $77.73$ & $76.67$ & 80.89 \\
PointPillars & 0   & $77.01$ & $74.77$ & $79.76$ \\
TANet      & 0    & $77.85$ & $75.62$ & $80.56$ \\
SECOND       & 0    &	78.43 &	$\mathbf{77.13}$&	$\mathbf{81.43}$\\
\textbf{SparseDet}     & 0   &	$\mathbf{78.45}$ &	77.03&	81.28\\ \midrule
PointRCNN     & 20  & $76.95$ & $74.73$ & $79.97$ \\
PointPillars & 20   & $76.74$ & $74.54$ & $79.50$ \\
TANet         & 20   & $77.68$ & $75.31$ & 80.39 \\
SECOND       & 20      &	78.19&	76.81&	81.17\\
\textbf{SparseDet }    & 20     &	$\mathbf{78.52}$ &	$\mathbf{76.93}$&	$\mathbf{81.33}$\\ \midrule
PointRCNN     & 100  & $75.98$ & $69.34$ & $77.63$ \\
PointPillars  & 100  & $76.06$ & $68.91$ & $77.20$ \\
TANet         & 100  & $76.64$ & $73.86$ & 79.34 \\
SECOND       & 100  &	77.27&	73.35&	79.7\\
\textbf{SparseDet }    & 100  &	\textbf{77.69} &	$\mathbf{74.44}$&	$\mathbf{80.12}$\\ \bottomrule
\end{tabularx}
\label{tab:robust}
\end{table}



In this section, we conduct extensive experiments to demonstrate the effectiveness of SparseDet and reveal its design concepts.
All the experiments are conducted upon the popular KITTI~\cite{geiger2012we} dataset. KITTI is a 3D object detection dataset for autonomous driving in the wild with $7481$ training samples.
Following F-PointNet~\cite{qi2018frustum}, we divide the training samples into a training set consisting of 3717 samples and a validation set that contains 3769 samples.
We evaluate models on the Car category with the official evaluation protocol where the IoU threshold is set as 0.7.
Mean average precision (mAP) with 11 recall positions under three difficulty levels (Easy, Moderate and Hard) is reported for a fair comparison with previous methods.

\subsection{Implementation Details}

{\bfseries Network Architecture.} 
The raw point cloud range is clipped into [0,70.4] m, [-40,40] m, and [-3,1] m for the XYZ axes according to the FOV annotation of KITTI Dataset. Then we adopt a voxel size of (0.05m, 0.05m, 0.1m) to discrete the whole point cloud into regular grids. 
Standard data-augmentation for point clouds is performed following the mainstream strategies in ~\cite{yan2018second,lang2018pointpillars,shi2019pointrcnn,shi2020points}, etc. Specifically, it includes randomly sampling ground truth objects from the database to join the current point cloud, randomly flipping points in a 3D bounding box along the X-axis, global rotation and scaling, etc. The maximum number of points $T$ in each voxel is set as 5.

The 3D sparse convolution backbone does one convolution block of its own resolution and three consecutive $2x$ down-sampling convolutions to obtain an $8x$ down-sampled feature map with filter numbers of (16, 32, 48, 64) respectively. 
After the feature alignment process, three convolution block with 4 layers in each is utilized to obtain a feature list consisting of its origin, half and quarter resolution of the input feature. The dimensions of these blocks are (64, 128, 256). Then the feature list is fed into the PSA-FPN feature extractor, transformed as a new same resolution feature list with the dimension of (128, 128, 128). As for the detection head, the proposal boxes number $N$ is set as 100, and the dimension of the proposal features is 128. A single head is stacked 6 times. Layers of the heads are initialized with Xavier~\cite{glorot2010understanding}. The pooling resolution $S$ is 7. We set $\lambda_{c l s}$, $\lambda_{L 1}$, and $\lambda_{IoU}$ to 2.0, 5.0, 2.0 respectively following~\cite{carion2020end,peize2020sparse}.

{\bfseries Training details.} 
The whole framework of SparseDet is optimized with the Adam optimizer in an end-to-end manner. All models are trained with 8 GPUs for 320 epochs with batch size 48, which takes around 6 hours. The learning rate is initialized as 0.003 and updated by a cosine annealing strategy for the learning rate decay.

{\bfseries Inference details.} 
At the inference process, the proposed boxes processed by the stacked heads will be directly used as the prediction result without any post-processing. The score of classification results shows the confidence of the prediction. 

\subsection{Comparison with SOTA}

{\bfseries Evaluation of 3D object detection on KITTI.}
We compare our SparseDet with several state-of-the-art methods on the 3D detection benchmark and the bird’s eye view detection benchmark in Tab.~\ref{tab:sota}. We achieved 81.28\% and 88.11\% average precision in the two tasks respectively with the real-time processing frame rate (34.5 FPS).

In terms of accuracy, SparseDet outperforms previous multi-sensor methods with large margins by only using LiDAR point clouds for the 3D detection benchmark. For both two benchmarks, SparseDet outperforms previous voxel-based methods, whose backbone is more similar to ours. At the same time, it is almost on par with the point-based methods. Which directly take raw point clouds as their operation inputs instead of converting a point cloud into a regular discrete representation. So the point-based methods are generally more accurate than voxel-based methods due to their fine-grained processing unit.

In terms of efficiency, our method is almost the fastest due to its sparse and end-to-end nature. It has an acceptable number of parameters and significantly fewer FLOPs. PointRCNN~\cite{shi2019pointrcnn} seems better in parameters and FLOPs, but the running time is relatively long due to their hard work of point-wise operations. By comparison, our SparseDet makes the best balance between effectiveness and efficiency among all the methods.

\begin{figure*}[t]
\begin{center}
  \includegraphics[width=0.96\linewidth]{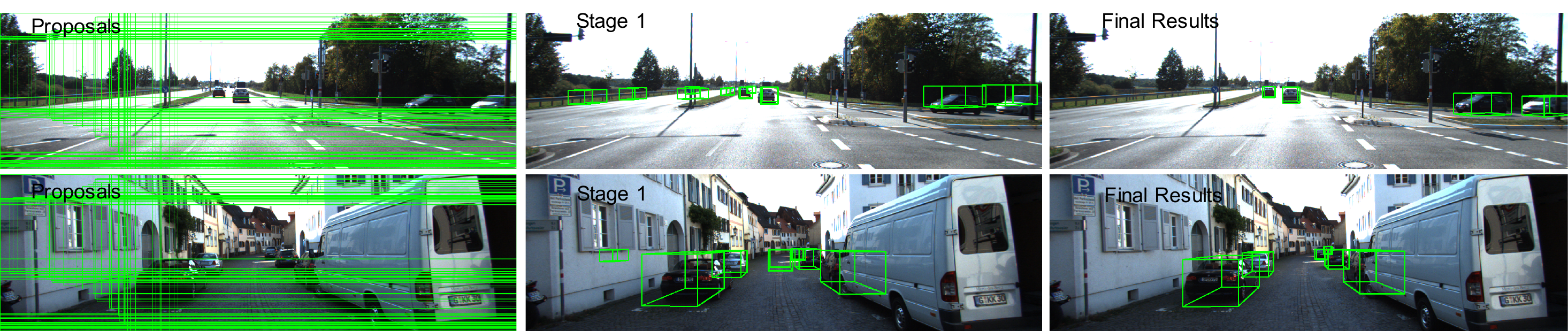}
\end{center}
\caption{The proposal boxes behavior visualization. At first, the learned proposals are randomly distributed in the point cloud space. Then they match to GTs gradually with the stacked heads structure. The effectiveness of sparse detection is proved.}
\label{behavior}
\end{figure*}

\begin{table}[!tp]
\centering
\caption{Ablation studies on feature alignment. To cooperate with the sparse detection, the feature aligner is re-devised.}
\begin{tabularx}{1.0\linewidth}{@{}l|*{5}X@{}} \toprule
Backbone & Easy & Mod. & Hard &  mAP \\ \midrule
Baseline & $83.38$ & $74.82$ & $73.48$ & $77.23$ \\
\textit{+PSA} & $84.29$ & $75.80$ & $73.93$ & $78.01$ \\
\textit{+FPN} & $84.65$ & $76.62$ & $75.38$ & $78.88$ \\ \midrule
Ours & $88.36$ & $78.45$ & $77.03$ & $81.28$\\ \bottomrule
\end{tabularx}
\label{tab:ablation-fe}
\end{table}

\begin{table}[t]
\centering
\caption{Impact of terms on our loss function. The integration of both $L1$ and $IoU$ loss into our formulation is crucial for the effectiveness of SparseDet. 
The details are given in Sec.~\ref{sec:ablation}
}.
\begin{tabularx}{1.0\linewidth}{*{3}X@{}|*{4}X@{}} \toprule
Class  & IoU & $\ell_{1}$ & Easy & Mod. & Hard & mAP   \\ \midrule
$\checkmark$ & $\checkmark$ &  $\times$ & 0.00&0.00&0.00&0.00 \\
\checkmark & $\times$ & $\checkmark$ &11.63&13.50&14.25& 13.13 \\
\checkmark & $\checkmark$ & $\checkmark$ & 88.36 & 78.45& 77.03&81.28\\
\bottomrule
\end{tabularx}
\label{tab:lossablat}
\end{table}

{\bfseries Solidness.} 
The performance of a detector will be greatly affected by the hyper-parameters under a single matching IoU threshold~\cite{smith2018improving}. So we utilize three different matching thresholds {0.70, 0.75, 0.80} for evaluation following~\cite{zhou2019IoU}. We compare the evaluation results with PointPillars~\cite{lang2018pointpillars} family on the 3D benchmark, and with PointRCNN~\cite{shi2019pointrcnn} family on the BEV benchmark in Tab.~\ref{tab:ap}. The results show that our method outperforms predecessors when the IoU threshold is high, especially in the BEV benchmark. It is proved that our method without any hand-crafted prior is solid. At the same time, BEV results are better than 3D results, which shows that the Z-axis dimension in 3D detection is difficult to match.

{\bfseries Robustness.} \label{sec:roubstness}
This method is more robust to noisy input data because its accuracy has less relevance to the input data variation due to its sparse property. Following TANet~\cite{liu2020tanet}, we introduce noise points to each object to verify robustness. As shown in Tab.~\ref{tab:robust}, we add three different numbers of noise points to the objects. The data of PointRCNN~\cite{shi2019pointrcnn}, PointPillars~\cite{lang2018pointpillars}, and TANet~\cite{liu2020tanet} comes from TANet~\cite{liu2020tanet} paper. While the data of SECOND~\cite{yan2018second} is reproduced by us using OpenPCDet~\cite{openpcdet2020}.


After the noise points are added, the accuracy of every method will decrease inevitably. Leaving aside the specific accuracy number, we only observe the declining trend of accuracy. In terms of hard category and overall mAP, SECOND~\cite{yan2018second} is better than SparseDet at the beginning, but with the increase of noise points, its accuracy is gradually surpassed by SparseDet. The results show that the average decline rates of mAP of the methods under discussion are 1.36\%, 1.08\%, 0.51\%, 0.71\%, and 0.48\% respectively. The decline speed of our SparseDet is significantly lower than other methods. TANet~\cite{liu2020tanet}, which is specially designed to solve the problem of robustness, has a close descending speed as ours because they add the point-wise, channel-wise, and voxel-wise attention mechanism in the voxel feature extraction. So that the information of the input point cloud can be more fully extracted. However, we do not perform such complex interactions, and the result is still better than TANet~\cite{liu2020tanet}, which proves the pretty robustness of our method.


\subsection{Ablation Studies} \label{sec:ablation}

\begin{table}[tp]
\centering
\caption{ Ablations on angle regression loss. Since the sparse detection is performed without guidance from anchors nor voxel correspondence, a proper loss is demanded to effectively supervise the angle regression.}
\begin{tabularx}{1.0\linewidth}{@{}l|*{5}X@{}} \toprule
$\mathcal{L}_{\theta}$  & Easy & Mod. & Hard & mAP & $\Delta$   \\
\hline $L1$ & 81.69 & 72.86 & 72.92 & 75.82	& $-5.46$ \\
$L1(sin,cos)$ & 83.41 & 74.49 & 71.57 & 76.49 & $-4.79$ \\ \midrule
$Sin-Error$ & 88.36 & 78.46 & 77.03 & 81.28	& - \\ \bottomrule
\end{tabularx}
\label{tab:thetalossablat}
\end{table}

\begin{table}[tp]
\centering
\caption{Ablations on 3D box matching constraints for the training of sparse prediction. $Cost$ stands for the matching cost, and $Loss$ stands for the training loss of matched pairs. $-\theta$ means discarding the angle constraint and $-h$ means discarding the height constraint. See Sec.~\ref{sec:loss} and ~\ref{sec:ablation} for details.}
\begin{tabularx}{1.0\linewidth}{ll|XXXX} \toprule 
Cost & Loss   & Mod. & Hard & mAP & $\Delta$   \\
\hline 3D&	3D&		0&	0&	0&		$-81.29$\\
(-h)&	(-h)		&59.15&	58.03&	60.55&		$-20.74$\\
(-$\theta$)&	(-$\theta$)&	76.91&	75.55&	79.24&		$-2.05$\\
(-$\theta$,-h)&	(-$\theta$,-h)&		77.24&	75.55&	79.64&	$-1.65$\\
(-$\theta$,-h)&	(-$\theta$)& 	75.12&	74.77&	76.77&		$-4.51$\\ \midrule
(-$\theta$,-h)&	3D&		78.48&	77.03&	81.29	&-	\\	\bottomrule
\end{tabularx}
\label{tab:ioulossablat}
\end{table}

Feature alignment fashion and loss design for the sparse detection head are the keys of SparseDet. In this section, we explore how the key component influence the final performance with extensive ablation experiments. All the ablation experiments are performed on the car class of the KITTI 3D benchmark.

{\bfseries Importance of PSA-FPN module.}
The widely used feature aligner in predecessors is a simple module in a conv-deconv way~\cite{yan2018second,lang2018pointpillars}, which is reproduced as a baseline. We remove the two parts of our novel feature aligner one by one as a comparative experiment to verify the effectiveness of the overall design.  The baseline achieves a 3D mAP of 77.23\%. With only PSA and only FPN structure, the performance is boosted to 78.01\% and 78.88\%, respectively. While the joint mechanism yields a 3D mAP of 81.28\%, outperforming the baseline model by 5.24\% and improving the accuracy of each category steadily. It is obvious that our feature aligner is more effective and robust. The detailed results of each difficulty category are shown in Tab.~\ref{tab:ablation-fe}.

{\bfseries Loss ablations.} There are three components of the overall loss: classification loss, $\ell_{1}$ bounding box distance loss, and IoU loss. We have executed extensive ablation experiments to study the importance of each component. First, we turn each component on and off to evaluate the overall effect of each part. Then we look into the specific part to study the effectiveness by conducting different alternatives. Specifically, the choice of $\theta$ loss and IoU loss of the matching cost and the loss are deliberated respectively.

On the overall effect study of each loss component, a model without the IoU loss and a model without the $L1$ distance loss are trained respectively to compare with the complete form. The corresponding results are presented in Tab.~\ref{tab:lossablat}. The model without $L1$ loss can not converge at all, leading to a complete failure on this task. While the model without IoU loss is a little better with 13.13\% mAP, but is still a failure. The surprisingly poor results of the two incomplete models show that each component is indispensable for 3D detection. And the $L1$ loss is at a more dominant position. These results are quite different from the situation of 2D images. Whose $L1$ loss and $IoU$ loss are a complement of each other, which would not lead to the failure of the task, but only a slight decline~\cite{carion2020end}. This phenomenon reveals that the 3D detection task has more challenges than 2D.
Moreover, previous 3D detectors almost have no IoU loss also show satisfying detection results. This phenomenon reveals that sparse 3D detection is more difficult than dense fashion.

Then, we study the details of $L1$ loss and IoU loss respectively. In terms of $L1$ loss, there are two parts including $(x,y,z,w,l,h)$ and {$\theta$} of the regression results as Equ.\ref{func:l1}. We calculate $L1$ distance directly with the first part because it is trivial without particularity. But as for the second part $\theta$, it has a great influence on the detection effect for its periodicity. As shown in Tab.~\ref{tab:thetalossablat}, the most direct way is to subtract the angle directly, it shows degradation of 6.72\% of our final Sin-Error choice. Another alternative to solve the periodicity is to convert the angle into a vector encoded by (sin, cos) pair~\cite{ku2018joint}. But in this way, two angles corresponding to the same box may generate a large loss for that each radian has a unique code, e.g. the 0 and $\pi$ radians. A 5.89\% degradation appeared using this loss. While the Sin-Error only uses the sin value of the angle difference, fixing the adversarial problem, which is more reasonable to constrain the angle regression.

In terms of $IoU$ loss, there are two stages of the matching cost and the matched loss. As stated in Set.\ref{sec:loss}, the direct use of rotated 3D $IoU$ cost and loss would not work at all. While our coarse-to-fine way consists of 2D $IoU$ loss without angle and 3D $IoU$ loss with angle exploiting a pretty good solution. To illustrate the effectiveness of this design, we show the comparison results by using different constraints in Tab.~\ref{tab:ioulossablat}. By alternatively discarding $\theta$, height, or the combination of them, we validate the success of our design. The results inspire us that we should relax the restrictions on some dimensions to achieve better results sometimes.

\subsection{Visualization}

 {\bfseries Qualitative Results.}
 Some qualitative results of our proposed SparseDet are shown in Fig.\ref{visualization}. Our SparseDet only takes the point cloud as input and the image is just for better visualization. The score threshold is set as 0.2 when drawing the prediction boxes.  We can find that the detector is fairly smart to discriminate the similar classes of $Truck$ and $Van$ from the predicted target $Car$ class.
 
 {\bfseries The Proposal Boxes Behavior.}
Fig.\ref{behavior} shows the behavior of learned proposal boxes. We project the 3D boxes onto the image for convenient visualization. In the beginning, the proposals are rather big boxes, including almost the whole point cloud in different orientations. After continuous optimization of the stacked heads, precise results are obtained. It demonstrates that acceptable results can be produced even at the first head, which shows the effectiveness of the sparse detection. Subsequent heads are used to refine the predictions gradually to generate the final results.

\section{\uppercase{Conclusions}}

In this paper, we present an end-to-end paradigm for 3D object detection in point cloud based on sparse prediction, namely SparseDet.
It maintains and iteratively refines a sparse set of proposals to directly produce final object detection without redundant removal and non-maximum suppression.
To collaborate with the sparse prediction, we closely investigate key components for effective 3D object detection and properly design the network.
The proposed method achieves competitive results in the KITTI dataset, with significant and consistent improvement over the baseline.
Moreover, the presented SparseDet is more robust to noisy data in point cloud.

SparseDet demonstrates the potential of end-to-end 3D object detection with sparse prediction. We hope it will inspire more exploration into the sparsity of 3D object detection and open up a new opportunity in this field. As one of the exploration directions, we will next try to combine the sparse detection paradigm with the point-based backbone to further improve the accuracy of the prediction results.

\section*{\uppercase{Acknowledgements}}

This work was supported by National Key Research and Development Program of China [grant number 2018YFB1403901] and National Natural Science Foundation of China (NSFC) [grant number 61872014].

\bibliographystyle{apalike}
{\small
\bibliography{example}}



\end{document}